\documentclass[accepted]{uai2025} 
                        
\usepackage[american]{babel}
\usepackage{hyperref}
\usepackage{url}

\usepackage{url}
\usepackage{amsfonts}
\usepackage{amsthm}
\usepackage[utf8]{inputenc}
\usepackage{graphicx}
\usepackage{colortbl}
\usepackage{tabulary}
\usepackage{subcaption}
\usepackage{pgf} 
\usepackage{graphics}
\usepackage{booktabs}       
\usepackage{multirow}
\usepackage{pgfplots}

\usepackage{hyperref}
\usepackage{etoolbox}
\usepackage{collcell}
\usepackage{nicefrac}       
\usepackage{microtype}      
\usepackage{natbib} 
\usepackage{xcolor}         
\usepackage[ruled,linesnumbered,noend]{algorithm2e}
\usepackage{pifont}
\usepackage{adjustbox}
\usepackage{makecell}
\usepackage{mathtools} 
\usepackage{wrapfig}

\newcommand{\indep}{\perp \!\!\! \perp}

\newtheorem{assum}{Assumption}

\newcommand{\SumNode}{\mathsf{S}}
\newcommand{\ClientNode}{\mathsf{C}}
\newcommand{\ProductNode}{\mathsf{P}}
\newcommand{\Node}{\mathsf{N}}
\newcommand{\Leaf}{\mathsf{L}}
\newcommand{\graph}{\mathcal{G}}
\newcommand{\ch}[1]{\operatorname{ch}(#1)}

\usepackage{xargs}
\usepackage[colorinlistoftodos,textsize=tiny]{todonotes} 
\newcommandx{\todoc}[2][1=]{{\todo[linecolor=orange,backgroundcolor=orange!25,bordercolor=orange,#1]{\tiny
      TODO: #2}}}
\newcommandx{\unsure}[2][1=]{{\todo[linecolor=yellow,backgroundcolor=yellow!25,bordercolor=yellow,#1]{\tiny
      UNSURE: #2}}}
\newcommandx{\change}[2][1=]{{\todo[linecolor=blue,backgroundcolor=blue!25,bordercolor=blue,#1]{\tiny
      CHANGE: #2}}}
\newcommandx{\info}[2][1=]{{\todo[linecolor=green,backgroundcolor=green!25,bordercolor=green,#1]{\tiny
      INFO: #2}}}
\newcommandx{\improvement}[2][1=]{{\todo[linecolor=violet,backgroundcolor=violet!25,bordercolor=violet,#1]{\tiny
      IMPROVEMENT: #2}}}
\newcommandx{\fb}[2][1=]{{\todo[inline,linecolor=lime,backgroundcolor=lime!25,bordercolor=lime,#1]{\tiny
      FB: #2}}}
\newcommandx{\thiswillnotshow}[2][1=]{{\todo[disable,#1]{THIS WILL NOT SHOW:
      #2}}}

\newtheorem{defin}{Definition}
\newtheorem{prop}{Proposition}

\usepackage{natbib} 
    \renewcommand{\bibsection}{\subsubsection*{References}}
\usepackage{mathtools} 
\usepackage{booktabs} 
\usepackage{tikz} 

\definecolor{qUmr}{HTML}{AB0392}
\definecolor{4iCr}{HTML}{3492EB}
\definecolor{6dQr}{HTML}{15A123}
\definecolor{allrev}{HTML}{f59342}

\title{Scaling Probabilistic Circuits via Data Partitioning}

\author[1]{\href{mailto:<jonas.seng@tu-darmstadt.de>?Subject=Federated Circuits}{Jonas Seng}{}}
\author[1,2]{Florian P. Busch}
\author[4]{Pooja Prasad}
\author[4]{Devendra S. Dhami}
\author[5]{Martin Mundt}
\author[1,2,3]{Kristian Kersting}
\affil[1]{%
    Computer Science Department, TU Darmstadt
}
\affil[2]{%
    Hessian Center for AI (hessian.AI)
}
\affil[3]{%
    German Research Center for AI (DFKI)
}
\affil[4]{
    Department of Mathematics and Computer Science, Eindhoven University of Technology
}
\affil[5]{
    Department of Mathematics and Computer Science, University of Bremen
}
  
\begin{document}
\maketitle

\begin{abstract}
Probabilistic circuits (PCs) enable us to learn joint distributions over a set of random variables and to perform various probabilistic queries in a tractable fashion. Though the tractability property allows PCs to scale beyond non-tractable models such as Bayesian Networks, scaling training and inference of PCs to larger, real-world datasets remains challenging. To remedy the situation, we show how PCs can be learned across multiple machines by recursively partitioning a distributed dataset, thereby unveiling a deep connection between PCs and federated learning (FL). This leads to federated circuits (FCs)---a novel and flexible federated learning (FL) framework that (1) allows one to scale PCs on distributed learning environments (2) train PCs faster and (3) unifies for the first time horizontal, vertical, and hybrid FL in one framework by re-framing FL as a density estimation problem over distributed datasets. We demonstrate FC's capability to scale PCs on various large-scale datasets. Also, we show FC's versatility in handling horizontal, vertical, and hybrid FL within a unified framework on multiple classification tasks.
\end{abstract}

\section{Introduction}

Probabilistic Circuits (PCs) are a family of models that provide tractable inference for various probabilistic queries~\citep{domingos2011spns, Choi2020ProbabilisticCA}.
This is achieved by representing a joint distribution by a computation graph on which certain structural properties are imposed.
While PCs offer significant computational advantages over traditional probabilistic models such as Bayesian networks~\citep{pearl1985bayesian}, further performance gains can be realized by optimizing the compactness of PC representations and tailoring them to specific hardware architectures~\citep{perharz2020einsum,liu2024scalingtractableprobabilisticcircuits}.
However, another natural way to scale up PCs by distributing the model over multiple machines is so far underexplored.
While models like neural networks can be partitioned over multiple machines with relatively low efforts, partitioning PCs is more challenging as they come with certain structural constraints to ensure the validity of the represented joint distribution. Interestingly, we find an inherent connection between the structure of PCs and the paradigm of federated learning (FL).
In PCs, sum nodes combine probability distributions over the same set of variables via a mixture.
This resembles the horizontal FL setting~\citep{konevcny2016federated,li2020federatedChallenges}, where all clients hold the same features but different samples.
In contrast, the case of vertical FL~\citep{yang2019federated, wu2020privacy} in which the same samples are shared, but features are split across clients, can be linked to the product nodes used in PCs, which combine distributions of a disjoint set of variables.
Consequently, the hybrid FL~\citep{Zhang2020HybridFLAlgosAndImplementations} setting, where both samples and features are separated across clients, can be represented by a combination of sum and product nodes.
Thus, PCs are well positioned to connect all three FL settings in a unified way -- an endeavor considered hard to achieve in the FL community~\citep{Li_2023FLSurvey, wen2023federated}.

\begin{figure*}[t]
    \centering
    \includegraphics[width=.9\textwidth]{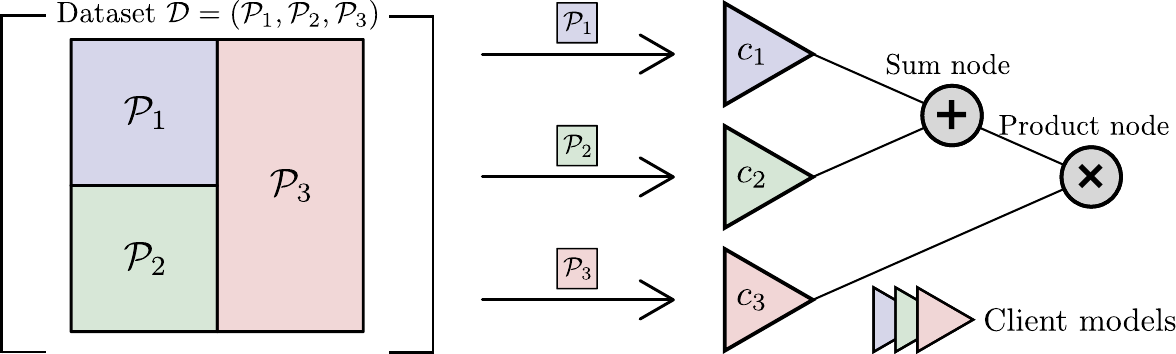}
    \caption{\textbf{Scaling PCs via Federated Circuits.} We scale PCs by splitting a dataset $\mathcal{D}$ into a set of $n$ partitions $\{\mathcal{P}_i\}_{i=1}^n$ s.t. $\mathcal{D} = \bigcup_{i=1}^n \mathcal{P}_i$. Each partition is assigned to a client (i.e., machine) $c_j$, and the resulting federated circuit (FC) is learned jointly by a set of clients. As a novel framework for federated learning (FL), FCs can perform horizontal FL (samples are split across clients), vertical FL (features are split across clients), and hybrid FL (mix of horizontal and vertical).}
    \label{fig:arch}
\end{figure*}

As a result of this connection, we introduce \textit{federated circuits (FCs)}, a novel FL framework that re-frames FL as a density estimation problem over a set of datasets distributed over multiple machines (subsequently called clients). 
FCs naturally handle all three FL settings and, therefore, provide a flexible way of scaling up PCs by learning a joint distribution over a dataset arbitrarily partitioned across a set of clients (see Fig. \ref{fig:arch} for an illustration).
Imposing the same structural properties as for PCs, FCs achieve tractable computation of probabilistic queries like marginalization and conditioning across multiple machines.
Based on this, we propose a highly communication-efficient learning algorithm that leverages the semi-ring structure within the design of FCs.
Our experimental evaluation shows that FCs outperform EiNets~\citep{perharz2020einsum} and PyJuice~\citep{liu2024scalingtractableprobabilisticcircuits} on large-scale density estimation tasks, demonstrating the benefits of scaling up PCs. Additionally, FCs outperform or achieve competing results on various classification tasks in all federated settings compared to state-of-the-art neural network-based and tree-based methods, demonstrating its effectiveness in FL.
We make the following contributions:
\textbf{(1)} We introduce FCs, a communication-efficient and scalable FL framework unifying horizontal, vertical, and hybrid FL by mapping the semantics of PCs to FL. \textbf{(2)}  We practically instantiate FCs to FedPCs and demonstrate how the FC framework can be leveraged to scale up PCs to large real-world datasets. \textbf{(3)} We propose a one-pass training scheme for FedPCs that is compatible with a broad set of learning algorithms. \textbf{(4)} We provide extensive experiments demonstrating the effectiveness of our approach for learning large-scale PCs and performing FL. We consider classification and density estimation on tabular and image data.

We proceed as follows: After touching upon related work, we provide the probabilistic view on FL and introduce FCs. Before concluding, we present our extensive experimental evaluation of FedPCs.
Our code is publicly available at \url{https://github.com/J0nasSeng/federated-spn.git}.

\section{Preliminaries and Related Work}
In the following, we briefly introduce PCs and FL and give an overview of relevant related work.

\textbf{Probabilistic Circuits.}
PCs encode a probability distribution as a computation graph that allows for tractable inference of a wide range of queries such as conditioning and marginalization. 
\cite{perharz2015theoreticalPC} define a PC over random variables $\mathbf{X}$ as a tuple $(\graph, \phi)$ where $\graph = (V, E)$ is a rooted, Directed Acyclic Graph (DAG) and $\phi: V \rightarrow 2^{\mathbf{X}}$ is the \textit{scope} function assigning a subset of random variables to each node in $\graph$. For each internal node $\Node$ of $\graph$, the scope is defined as $\phi(\Node) = \cup_{\Node' \in \ch{\Node}} \phi(\Node')$. Each leaf node $\Leaf$ computes a distribution/density over its scope. All internal nodes of $\graph$ are either a sum node $\SumNode$ or a product node $\ProductNode$ where each sum node computes a convex combination of its children, i.e. $\SumNode = \sum_{\Node \in \ch{\SumNode}} w_{\SumNode, \Node}\Node$, and each product node computes a product of its children, i.e. $\ProductNode = \prod_{\Node \in \ch{\ProductNode}}\Node$.
To ensure tractability,
a PC must be \textit{decomposable}. Decomposability requires that for all $\ProductNode \in V$ it holds that $\phi(\Node) \cap \phi(\Node') = \emptyset$ where $\Node, \Node' \in \ch{\ProductNode}$. To further ensure that a PC represents a valid distribution, \textit{smoothness} must hold, i.e., for each sum $\SumNode \in V$ it holds that $\phi(\Node) = \phi(\Node')$ where $\Node, \Node' \in \ch{\SumNode}$~\citep{domingos2011spns,peharz2015theoretical,sanchez2021sum}. 

Several works have tackled the goal of scaling PCs.
\citet{peharz2020random} have shown that learned PC structures can be replaced by large, random structures to scale to larger problems.
Changes in the model layout, such as parallelizable layers via einsum-operations~\citep{perharz2020einsum} and a reduction in IO operations~\citep{liu2024scalingtractableprobabilisticcircuits}, were also shown to reduce the speed of computation drastically.
\cite{lius2022calingviaLV} improved the performance of PCs by latent variable distillation using deep generative models for additional supervision during learning.

\textbf{Federated Learning.}
In federated learning (FL), a set of data owners (or clients) aim to collaboratively learn an ML model without sharing their data. One distinguishes between horizontal, vertical, and hybrid FL depending on how data is partitioned. In horizontal FL, a dataset $\mathbf{D} \in \mathbb{R}^{n \times d}$ is partitioned such that each client holds the same $d$ features but different, non-overlapping sets of samples. In vertical FL, $\mathbf{D}$ is partitioned such that each client holds the same $n$ samples but different, non-overlapping subsets of the $d$ features. Hybrid FL describes a combination of horizontal and vertical FL where clients can hold both different (but possibly overlapping) sets of samples and features~\citep{wen2023federated, Li_2023FLSurvey}.

For all three FL settings, specifically tailored methods have been proposed to enable collaborative learning of models.
The most common scheme in horizontal FL is to average the models of all clients regularly during training \citep{McMahan2016FedAvg, karimireddy2020mime, Karimireddy2020SCAFFOLD, Sahu2018FedProx}. However, model averaging requires each client to share the same model structure. In vertical FL, clients hold different feature sets; thus, there is no guarantee that the model structure can be shared among clients. In these cases, tree-based and neural models are the predominant choice and are typically learned by sharing data statistics or feature representations among clients \citep{Kourtellis2016VHT, Cheng2021Secureboost, vepakomma2018split,Cellabos2020SplitNN, Tianyi2020VAFL, Liu2019FedBCD}. 
Similar to tree-based vertical FL, tree-based hybrid FL approaches share data statistics (such as histograms) or model properties (such as split rules) among clients \citep{li2023fedtree, li2024effective}. However, tree-based approaches often require complex training procedures.


\section{Federated Circuits}
This work aims to scale up PCs by splitting data and the model across multiple machines, thus harnessing the availability of compute clusters to train PCs in a federated fashion.
In the following, we present an elegant and effective way to achieve that using our novel federated learning framework called federated circuits (FCs) that unifies horizontal, vertical, and hybrid FL.
\subsection{Problem Statement \& Modeling Assumptions}
Given a dataset $\mathbf{D}$ and a set of clients $\mathcal{C}$ where each $c \in \mathcal{C}$ holds a partition $\mathbf{D}_c$ of $\mathbf{D}$; we aim to learn the joint distribution $p(\mathbf{X})$ over random variables $\mathbf{X}$ (i.e., the features of $\mathbf{D}$).
The partitioning of $\mathbf{D}$ is not further specified. Hence, each client might only hold a subset of random variables $\mathbf{X}_c \subseteq \mathbf{X}$ with support $\mathcal{X}_c$. This can be interpreted as each $c \in \mathcal{C}$ holding a dataset $\mathbf{D}_c \sim p_c$ where $p_c$ is a joint distribution over $\mathbf{X}_c$ which is related to $p(\mathbf{X})$.
We introduce two critical modeling assumptions relevant for learning a joint distribution $p(\mathbf{X})$ from a dataset $\mathbf{D}$ partitioned across a set of machines.


\begin{assum}[Mixture Marginals]\label{assum:decomposition}
    There exists a joint distribution $p$ such that the relation $\int_{\mathbf{X} \setminus \mathbf{X}_S} p(x) = \sum_{l \in L} q(L=l) \cdot p_{S}(x | L=l)$ holds for all $x \in \mathcal{X}$. Here, $\mathbf{X}_{S} \subseteq \mathbf{X}$ is a subset of the union of client random variables $\mathbf{X} = \cup_{c \in \mathcal{C}} \mathbf{X}_c$. Further, $\mathcal{X} = \bigtimes_{c \in \mathcal{C}} \mathcal{X}_c$ is the support of $\mathbf{X}$, each $p_{S}$ is defined over $\mathbf{X}_S \subseteq \mathbf{X}$ and $q$ is a prior over a latent $L$.
\end{assum}

To illustrate, consider a subset of variables $\mathbf{X}_S \subseteq \mathbf{X}$ shared among all clients and its complement $\mathbf{X}_{S^-} = \mathbf{X} \setminus \mathbf{X}_S$.
Assumption \ref{assum:decomposition} ensures that the marginal $\int_{\mathbf{X}_{S^-}} p(\mathbf{X})$ is representable as a mixture of all client distributions $p_c(\mathbf{X}_S)$ over $\mathbf{X}_S$.
If Assumption \ref{assum:decomposition} would not hold, the information stored on the clients' data partitions would not be sufficient to learn $p(\mathbf{X})$.
A key assumption in FL is that data cannot be exchanged among clients. However, dependencies among variables residing on different clients might still exist. 
To enable learning these ``hidden" dependencies while keeping data private, we make the following assumption:

\begin{assum}[Cluster Independence]\label{assum:cluster_independence}
    Given disjoint sets of random variables $\mathbf{X}_1, \cdots, \mathbf{X}_n$ and a joint distribution $p(\mathbf{X}_1, \cdots, \mathbf{X}_n)$, assume that a latent $L$ can be introduced s.t. the joint can be represented as $p(\mathbf{X}_1, \cdots, \mathbf{X}_n) = \sum_l q(L=l) \prod_{i=1}^n p(\mathbf{X}_i | L=l)$ where $q$ is a prior distribution over the latent $L$.
\end{assum}

Note that independence is only assumed within clusters in the data. Thus, the latent variable (which can be thought of as "cluster selectors``) allows capturing dependencies among variables residing on different clients. 
Distributions of the form in Assumption \ref{assum:cluster_independence} are strictly more expressive than the product distribution and allow more complex modeling.

Further, we want to emphasize that Assumptions \ref{assum:decomposition} and \ref{assum:cluster_independence} are common throughout PC literature. Assumption \ref{assum:decomposition} forms the basis of the validity of marginalization, and Assumption \ref{assum:cluster_independence} plays a crucial role in constructing structure learning algorithms for PCs. For more details, refer to App. \ref{app:assumptions}.

\subsection{Bridging Probabilistic Circuits and Federated Learning}
\label{subsec:fcs}

We now illustrate an inherent connection between PC semantics and FL, allowing us to scale PCs on large datasets by partitioning the data over a set of clients.

\textbf{Sum Nodes and Horizontal FL.} In horizontal FL, each client is assumed to hold the same set of features, i.e., $\mathbf{X}_c = \mathbf{X}_{c'}$ for all $c, c' \in \mathcal{C}$. However, each client holds different samples. 
Prominent horizontal FL methods aggregate the \textit{model parameters} of locally learned models regularly during training.
However, the horizontal FL setting also precisely corresponds to the interpretation of sum nodes in PCs: A sum node splits a dataset into multiple disjoint clusters. This results in a mixture distribution representing the data that is learned from the disjoint clusters. 
Thus, instead of aggregating model parameters, we aggregate the \textit{distributions} learned by each client on its data partition.

\begin{defin}[Horizontal FL]\label{def:HFL}
    Assume a set of samples $\mathbf{D}_c \sim p_c$ on each client $c \in \mathcal{C}$, a joint distribution $p$ adhering to Assumption \ref{assum:decomposition} and that $\mathbf{X}_c = \mathbf{X}_{c'}$ for all $c, c' \in \mathcal{C}$ s.t. $c \neq c'$. We define horizontal FL as fitting a mixture distribution $\hat{p} = \sum_{c \in \mathcal{C}} q(c) \cdot \hat{p}_c$ such that $d(\hat{p}, p)$ and $d(p_c, \hat{p}_c)$ are minimal for all $c \in \mathcal{C}$ where $d$ is a distance metric and $\hat{p}_c$ local distribution estimates.
\end{defin}

This view on horizontal FL has an appealing positive side effect: Aggregating model parameters can lead to divergence during training if the client's data distributions significantly differ. Since we aggregate distributions in mixtures, we naturally can handle heterogeneous client distributions. Also, since clients can train models independently, the communication cost of the training is minimized.

\textbf{Product Nodes \& Vertical FL.} In vertical FL, each client is assumed to hold a disjoint set of features, i.e., $\mathbf{X}_c \cap \mathbf{X}_{c'} = \emptyset$ for all $c, c' \in \mathcal{C}$. In contrast to horizontal FL, all clients hold different features belonging to the same sample instances.
As in horizontal FL, there is a semantic connection between vertical FL and PCs. Product nodes in PCs compute a product distribution defined on a disjoint set of random variables. Thus, a product node separates the data along the feature dimension, corresponding to the vertical FL setting. However, a product node assumes the random variables of the child distributions to be independent of each other. Obviously, this is an unrealistic assumption for vertical FL, where features held by different clients might be statistically dependent. Assumption \ref{assum:cluster_independence} can be exploited to capture such dependencies, and a mixture of products of independent clusters can be formed. See Sec. \ref{subsec:FedPCs} for details.

\begin{defin}[Vertical FL]\label{def:VFL}
    Assume a set of samples $\mathbf{D}_c \sim p_c$ on each data owner $c \in \mathcal{C}$, the existence of a joint distribution $p$ adhering to Assumptions \ref{assum:decomposition} and \ref{assum:cluster_independence} and that $\mathbf{X}_c \cap \mathbf{X}_{c'} = \emptyset$ holds for all $c, c' \in \mathcal{C}$ s.t. $c \neq c'$. We define vertical FL as estimating a joint distribution $\hat{p}$ s.t. $d(p, \hat{p})$ is minimal and $\int_{\mathbf{X} \setminus \mathbf{X}_c} \hat{p}(x) = \hat{p}_{c}(x)$ for all $x \in \mathcal{X}$ where $d$ is a distance metric and $\hat{p}_c$ are estimates of client distributions.
\end{defin}

\textbf{PCs \& Hybrid FL.}
Given Defs. \ref{def:HFL} and \ref{def:VFL}, hybrid FL is a combination of both. 
In terms of PC semantics, this amounts to building a hierarchy of fusing marginals and learning mixtures.
Provided with these probabilistic semantics, we can now formally bridge PCs and FL.
In the following, we distinguish between clients $\mathcal{C}$ and servers $\mathcal{S}$ and define the set of machines participating in training as $\mathcal{N} = \mathcal{C} \cup \mathcal{S}$. Bringing everything together and abstracting from the probabilistic interpretation, we define \textbf{federated circuits} (FCs) as follows.

\begin{defin}[Federated Circuits]\label{def:fcs}
A \textbf{federated circuit} (FC) is a tuple $(\graph, \psi_{\graph}, \omega)$ where $\graph = (V, E)$ is a rooted, Directed Acyclic Graph (DAG), $\psi_{\graph}: V \rightarrow \mathcal{N}$ assigns each $\Node \in V$ to a client/server $n \in \mathcal{N}$ based on the structure of $\graph$ and $\omega: V \rightarrow O$ assigns an operation $o \in O$ to each node $\Node \in V$ where $o: \text{dom}(\ch{\Node}) \rightarrow \text{dom}(\Node)$ computes the value of $\Node$ given the values of the children of $\Node$.
\end{defin}

FCs extend the definition of PCs in the sense that FCs represent a 
computational graph $\mathcal{G} = (V, E)$ distributed over multiple machines where arbitrary operations can be performed in each node $\Node \in V$. Note that through $\mathcal{G}$, FCs also define the structure of a communication network among participating machines. Also, FCs are not restricted to the probabilistic interpretation presented above. For example, parameterizing leaves by decision trees and introducing a node $\Node$ that performs averaging yields a bagging model.
\subsection{Federated Probabilistic Circuits}
\label{subsec:FedPCs}
Let us now dive deeper into the probabilistic interpretation of FCs. To that end, we present a concrete instantiation of FCs leveraging Probabilistic Circuits (PCs) as leaf models, resulting in federated PCs (FedPCs).
Following the probabilistic interpretation from Sec. \ref{subsec:fcs}, we align the PC structure with the communication network structure to form a federated PC.
\begin{defin}[Federated PC]
A Federated PC (FedPC) is a FC where each leaf node $\ClientNode$ is a density estimator and each node $\Node$ s.t. $\ch{\Node} \neq \emptyset$ is either a sum node ($\SumNode$) or a product node ($\ProductNode$).
\end{defin}
Note that only the client nodes $\ClientNode$ hold a dataset and we only demand the clients to be parameterized by a density estimator.
In order for FedPCs to be computationally efficient, these density estimators should be tractable.
In the following, we parameterize the leaf nodes $\ClientNode$ as PCs.

The assignment function $\psi$ transforms the PC's computation graph into a distributed computation graph, thus inducing a communication network. 
This establishes a direct correspondence between PC semantics (computation graph) and the communication network structure in FedPCs. Inference is performed as usual in PCs by propagating likelihood values from the leaf nodes to the root node. The only difference is that the result of a node $\Node$ has to be sent to its parent(s) $\mathbf{pa}(\Node)$ over the communication network if $\psi(\Node) \neq \psi(\Node')$ holds for $\Node' \in \mathbf{pa}(\Node)$.

Training FedPCs requires adapting the regular training procedure for PCs because in FL, clients cannot access other clients' data. 
For example, training with Expectation Maximization (EM) requires access to the same samples for all clients, which is incompatible with horizontal and hybrid FL.
Similarly, LearnSPN~\cite {gens2013LearnSPN} requires access to all features due to independence tests performed during training.
To solve this, we propose a \textit{one-pass} training procedure for FedPCs.

\begin{algorithm}[t]
\caption{One-Pass Training}\label{alg:fedpcLearning}
\KwData{Clients $\mathcal{C}$, features $\mathbf{X}$, cluster size $K$, FedPC}
\KwResult{Trained fedPC}
Set $\mathbf{M} = \mathbf{0}^{|\mathcal{C}| \times |\mathbf{X}|}$ and map $= []$\;
$\mathbf{M}_{i, j} = 1$ if $X^{(j)}$ on client $i$\;
\For{$j, \mathbf{u}$ in enum. of distinct columns $\mathcal{U}$}{
    $\mathbf{S}^{(j)} = \{i : i \in \{1, \dots, |\mathbf{X}| \land \text{all}(\mathbf{u} == \mathbf{M}_{:, i})\} \}$\;
    $O_{\mathbf{S}^{(j)}} = \text{argwhere}(\mathbf{u} == 1)$\;
    map.append($\mathbf{S}^{(j)}$, $O_{\mathbf{S}^{(j)}}$)\;
}
sums $= []$\;
\For{$\mathbf{S}^{(j)}$, $O_{\mathbf{S}^{(j)}}$ in map}{
    \If{$|O_{\mathbf{S}^{(j)}}| > 1$}{
        s $=$ fedPC.add\_sum($\mathbf{S}^{(j)}$, $O_{\mathbf{S}^{(j)}}$)\;
        sums.add(s)
    }
    \Else{
        client\_clusters $=$ cluster\_local\_data($O_{\mathbf{S}^{(j)}}$, $K$)\;
    }
}
products $=$ fedPC.add\_products($P$)\;
\For{prod in products}{
    prod.children.add(sums)\;
    \For{client, clusters in client\_clusters}{
        prod.children.add\_rand\_subset(clusters)\;
    }
}
fedPC.add\_mixture\_over\_products(products)\;
fedPC.train\_clients()\;
fedPC.infer\_weights()\;
\Return fedPC
\end{algorithm}

\begin{figure*}
    \centering
    \includegraphics[width=.8\textwidth]{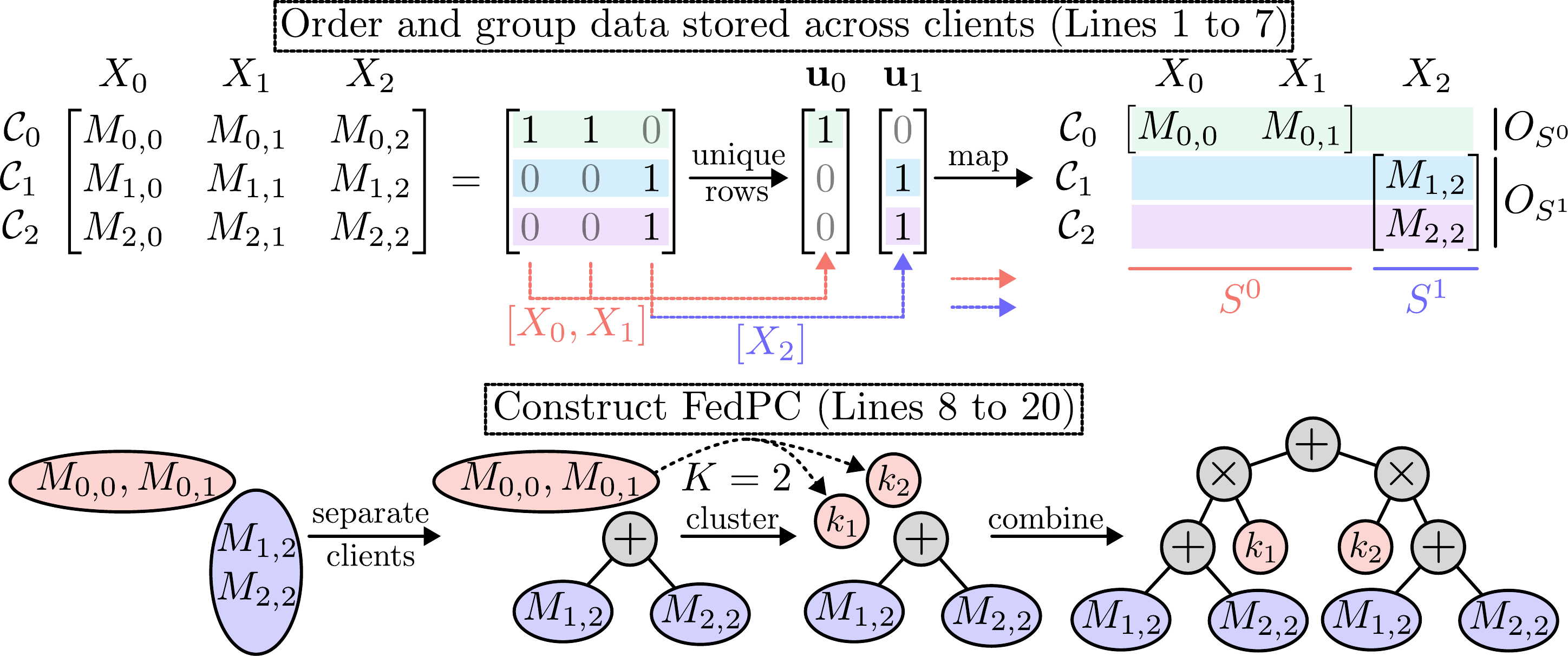}
    \caption{\textbf{One-Pass Training Visualized.} (Top) First, the matrix $\mathbf{M}$ is initialized, representing which features are held by which client. Feature subsets are constructed by considering distinct column vectors $\mathbf{u}$ of $\mathbf{M}$ that represent the same set of clients. This forms a mapping indicating which features are modeled as a mixture over clients. (Bottom) This mapping is utilized by forming mixtures over different clients sharing the same feature set via sum nodes. Features that are not shared over multiple clients will be clustered into $K$ clusters (here $K=2$). The FedPC is formed by creating product nodes containing all sum nodes from the previous steps and at least one of the $K$ clusters. Lastly, the root node is inserted.}
    \label{fig:algo}
\end{figure*}

\textbf{One-Pass Training.} Our one-pass learning algorithm learns the structure and parameters of FedPCs such that local models can be trained independently (Algo. \ref{alg:fedpcLearning}, Fig.~\ref{fig:algo}).
Before training, all clients $c \in \mathcal{C}$ share their set of uniquely identifiable features/random variables $\mathbf{X}_c$ with a server, resulting in the feature set indicator matrix $\mathbf{M}^{|\mathcal{C}| \times |\mathbf{X}|}$ \textbf{(Lines 1-2)}. Feature identifiers can be names of features such as ``account balance" and must correspond to the same random variable on all clients (thus uniquely identifiable). Then, the server divides the joint feature space $\mathbf{X}$ into disjoint subspaces $\mathbf{S}^{(j)}$. For this, we consider the set of distinct column vectors $\mathcal{U}$ of $\mathbf{M}$ where we denote distinct vectors as $\mathbf{u}$. Since each column of $\mathbf{M}$ indicates the set of clients a feature resides on, we can use each $\mathbf{u} \in \mathcal{U}$ to compute a set of features that are shared across the same set of clients. This results in $|\mathcal{U}|$ distinct feature sets, denoted $\{\mathbf{S}^{(1)}, \dots, \mathbf{S}^{(|\mathcal{U}|)}\}$. Each $O_{\mathbf{S}^{(j)}}$ denotes the set of clients that hold the features in $\mathbf{S}^{(j)}$. \textbf{(Lines 3-7)}. This procedure is illustrated in Fig. \ref{fig:algo} (top).

Afterward, the FedPC structure is constructed as shown in Fig.~\ref{fig:algo} (bottom): First, we build a mixture (sum node) for each subspace $\mathbf{S}^{(j)}$ where $|O_{\mathbf{S}^{(j)}}| > 1$, i.e., more than one client holds $\mathbf{S}^{(j)}$ \textbf{(Lines 9-12)}. This enables each client to learn a PC over $\mathbf{S}^{(j)}$ independently. After that,$|O_{\mathbf{S}^{(j)}}| = 1$ holds for all remaining $\mathbf{S}^{(j)}$. Also, the scope of the sums nodes introduced in the FedPC share no features with any of the remaining $\mathbf{S}^{(j)}$ 
since the server divided the feature space into disjoint subspaces. Therefore, we introduce $P$ product nodes to construct the remaining part of the FedPC.
To this end, we divide the data of all subspaces $\mathbf{S}^{(j)}$ where $|O_{\mathbf{S}^{(j)}}| = 1$ holds into $K$ clusters \textbf{(Line 14)}. Each client learns a dedicated PC for each cluster. To ensure that the FedPC spans the 
entire feature space of the clients, the children of product nodes are set as follows: Each sum node introduced 
in the FedPC becomes a child of each product node. Additionally, for each $\mathbf{S}^{(j)}$ where $|O_{\mathbf{S}^{(j)}}| = 1$ holds, we randomly select a PC learned over one of the $K$ clusters s.t. the scope of each product node spans $\mathbf{X}$, and each PC representing a cluster is the child of at least one product node.
Then, we build a mixture over all product nodes using a sum node \textbf{(Lines 15-20)}. Note that we seek to construct product nodes over independent clusters, which aligns with the maximum entropy principle (see App. \ref{app:max_entropy} for details).
Once the FedPC is constructed, all client-sided PCs are learned. Since clients learn their PCs independently, each client can use an arbitrary learning algorithm (even different ones).
As a last step, the network-sided parameters, i.e., the weights of network-sided sum nodes, of the FedPC are inferred \textbf{(Line 21-22)}. For each sum node $\SumNode$, the weight $\mathbf{w}_{\SumNode}^{(i)}$ associated with the $i$-th child (i.e., distribution) of $\SumNode$ is set to $\frac{\rho(\Node_i)}{\sum_i \rho(\Node_i)}$. Here, $\rho(\Node_i) = \sum_{\ClientNode \in \ch{\Node_i}} |\mathbf{D}_{\ClientNode}|$ where $\mathbf{D}_{\ClientNode}$ is the dataset used to train the leaf $\ClientNode$. Hence, the network-sided weights can be inferred without any forward or backward pass. 
Note that this approach reduces horizontal FL to learning a mixture of the client's data distributions and vertical FL to learning a mixture over $P$ product nodes.


\subsection{Analysis of Communication Efficiency}\label{subsec:CommunicationAnalysis}
As a key requirement for efficient training when learning models at scale on partitioned data, we now analyze the communication efficiency of FedPCs. 

\textbf{Horizontal FL.}
Assume a client set $\mathcal{C}$ where each client holds a model with $M$ parameters. Further, assume models are aggregated $K$ times during training ($K$ communication rounds). Then, model aggregation-based algorithms like FedAvg commonly used in horizontal FL send $\mathcal{O}(M \cdot |\mathcal{C}| \cdot K)$ messages over the network as each client sends $M$ model parameters to a server in each communication round. Training FedPCs with one-pass training, in contrast, only requires $\mathcal{O}(|\mathcal{C}| \cdot (M + 1))$ messages over the network as models are learned locally and independently, followed by setting the parameters ($\mathcal{O}(|\mathcal{C}|)$ messages) of the sum nodes and aggregating the model on the server ($\mathcal{O}(M |\mathcal{C}|)$ messages).

\textbf{Vertical FL.}
In vertical settings, SplitNN-like architectures are commonly used. Assume training a SplitNN architecture for $E$ epochs that output a feature vector of size $F$ for each sample of a dataset with $S$ samples, vertically distributed over clients $\mathcal{C}$. The training requires sending $\mathcal{O}(E \cdot |\mathcal{C}| \cdot F \cdot S)$ messages over the network. In contrast, with one-pass training of FedPCs, each client learns a dedicated PC with $M$ parameters for each of the $K$ clusters that are learned. The last layer of the FedPC is a mixture of $P$ products of clusters. The mixture parameters are set after training each client's model. 
Aggregating the learned models and setting the network-sided mixture parameters requires $\mathcal{O}(K \cdot M \cdot |\mathcal{C}| + P)$ messages to be sent. 
If $(K \cdot M + \frac{P}{|\mathcal{C}|}) < (E \cdot F \cdot S)$ holds, training FedPCs is more communication efficient than training SplitNN-like architectures. In practice, this is likely to hold: The number of clusters is usually smaller than $100$ while feature vectors can have hundreds of dimensions (i.e., $F > 100$). Further, models should have fewer parameters than samples in the dataset to ensure generalization (i.e., $M < S$). $P$ can be set to an arbitrary value, depending on $|\mathcal{C}|$ and the data.
App. \ref{app:commeff} provides more details and an intuition on communication costs.

\textbf{Hybrid FL.}
In hybrid FL, FedPCs are trained on several subspaces: Some exist on all or a subset of clients (denoted as $R_s$) and some are only available on one client (denoted as $R_d$). Further denote communication costs of FedPCs in horizontal FL and vertical FL as $C_h$ and $C_v$, respectively. Since the training procedure in hybrid cases essentially performs horizontal FL on shared feature spaces and vertical FL on disjoint feature spaces, $\mathcal{O}(|R_s| \cdot C_h + |R_v| \cdot C_v)$ messages are sent over the network during training.



\section{Experiments}

\begin{table*}[]
\resizebox{\textwidth}{!}{
\centering
\begin{tabular}{c|cccc|cccc}
       & \multicolumn{4}{c}{Log-Likelihood}    & \multicolumn{4}{c}{Relative Runtime} \\
       & cent.              & horizontal              & vertical    & hybrid              & cent. & horizontal       & vertical              & hybrid    \\ \hline
MNIST  & $3352$\scriptsize{$\pm 3.5$}  & $3350$\scriptsize{$\pm 3.2$}  & $3351$\scriptsize{$\pm 3.8$} & $3349$\scriptsize{$\pm 3.7$} & $1.0$ & $\mathbf{0.07}$\scriptsize{$\pm \mathbf{0.01}$} & $0.13$\scriptsize{$\pm 0.01$}  & $0.13$\scriptsize{$\pm 0.02$} \\
Income & $-11.5$\scriptsize{$\pm 0.1$} & $-11.4$\scriptsize{$\pm 3.5$} & $-11.9$\scriptsize{$\pm 3.3$} & $-12.0$\scriptsize{$\pm 1.5$} & $1.0$ & $\mathbf{0.17}$\scriptsize{$\pm \mathbf{0.02}$} & $0.236$\scriptsize{$\pm 0.01$} & $0.21$\scriptsize{$\pm 0.02$} \\
Cancer & $-38.9$\scriptsize{$\pm 0.3$} & $-38.5$\scriptsize{$\pm 1.1$} & $-38.6$\scriptsize{$\pm 0.5$} & $-38.7$\scriptsize{$\pm 1.5$} & $1.0$ & $\mathbf{0.21}$\scriptsize{$\pm \mathbf{0.07}$} & $0.35$\scriptsize{$\pm 0.05$} & $0.35$\scriptsize{$\pm 0.1$}  \\
Credit & $-12.8$\scriptsize{$\pm 1.0$} & $-13.1$\scriptsize{$\pm 0.5$} & $-12.5$\scriptsize{$\pm 2.3$} & $-12.5$\scriptsize{$\pm 1.3$} & $1.0$ & $0.42$\scriptsize{$\pm 0.05$} & $\mathbf{0.31}$\scriptsize{$\pm 0.09$} & $0.40$\scriptsize{$\pm 0.13$}
\end{tabular}
}
\caption{\textbf{FedPCs speed up training while retaining model performance.} We trained PCs in a centralized setting (cent.) and in all FL settings (using FedPCs) on different datasets and the same structure learning algorithm. We find that FedPCs tremendously speed up training while there is no reduction in log-likelihood. This demonstrates that PCs can be learned in federated settings (for MNIST, log densities are reported). We report relative runtime where centralized runtime is 1.0.}
\label{tab:likleihoods}
\vspace{-0.25cm}
\end{table*}

\begin{figure*}[t]
    \begin{minipage}{0.48\textwidth}
        \includegraphics[width=.85\textwidth]{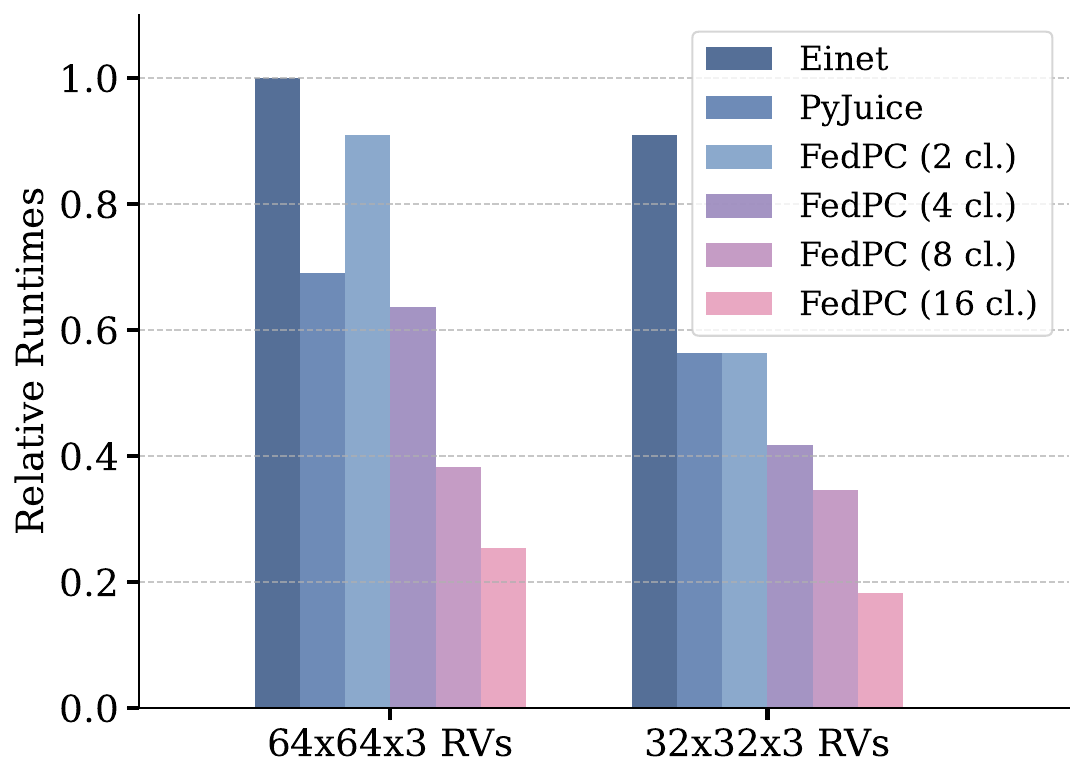}
        \caption{\textbf{FedPCs speed up training} on large-scale image data (64x64 and 32x32 RGB images) due to parallel training on separate data partitions.}
        \label{fig:runtime}
    \end{minipage}
    \hfill
    \begin{minipage}{.48\textwidth}
        \resizebox{\columnwidth}{!}{%
            \begin{tabular}{l|lll}
                              & CelebA             & Imagenet32          & Imagenet              \\ \hline
            EiNet              & -3.42 \scriptsize{$\pm$ 0.06} & -3.71 $\pm$ \scriptsize{0.04}   & -3.73 $\pm$ \scriptsize{0.04}  \\
            PyJuice           & -2.98 \scriptsize{$\pm$ 0.02} & -3.60 $\pm$ \scriptsize{0.01}  & -3.43 $\pm$ \scriptsize{0.02}   \\
            FedPC (2 cl.)  & -2.87 \scriptsize{$\pm$ 0.05} & -2.66 $\pm$ \scriptsize{0.02}  & -3.12 $\pm$ \scriptsize{0.02} \\
            FedPC (4 cl.)  & -2.84 \scriptsize{$\pm$ 0.05} & -2.56 $\pm$ \scriptsize{0.03}  & -3.01 $\pm$ \scriptsize{0.03} \\
            FedPC (8 cl.)  & \underline{-2.76} \scriptsize{$\pm$ \underline{0.04}} & \underline{-2.50} $\pm$ \underline{\scriptsize{0.03}}  & \underline{-2.97} $\pm$ \underline{\scriptsize{0.02}}  \\
            FedPC (16 cl.) & \textbf{-2.68} \scriptsize{$\pm$ \textbf{0.03}} & \textbf{-2.45} $\pm$ \textbf{\scriptsize{0.04}} & \textbf{-2.90} $\pm$ \textbf{\scriptsize{0.03}}
            \end{tabular}
            }
            \captionof{table}{\textbf{FedPCs outperform EiNets and PyJuice on density estimation tasks.} FedPCs achieve better results on density estimation tasks on tCelebA, Imagenet32, and Imagenet because they can learn large models distributed across multiple machines. Results reported in nats (higher is better). Best in \textbf{bold}, 2nd best \underline{underlined}.}
            \label{tab:imageLLs}
    \end{minipage}
    \vspace{-0.5cm}
\end{figure*}

Our empirical evaluation corroborates that FedPCs can be leveraged to scale up PCs effectively via data and model
partitioning. By performing horizontal, vertical and hybrid FL in one unified framework, we obtain high-performing models with the same or improved performance compared to prominent FL baselines. 

We aim to answer the following questions: \textbf{(Q1)}~Can FedPCs decrease the required training time and successfully learn a joint distribution over distributed data? \textbf{(Q2)}~Do FedPCs effectively scale up PCs, thus yielding more expressive models?
\textbf{(Q3)}~How do FCs with different parameterizations perform on classification tasks compared to existing FL methods? \textbf{(Q4)}~How does our one-pass learning algorithm compare to training with the EM algorithm?


\textbf{Experimental Setup.} To see if FedPCs, an instantiation of FCs, successfully scale up PCs, we follow \citet{liu2024scalingtractableprobabilisticcircuits} and perform density estimation on three large-scale, high-resolution image datasets:  Imagenet, Imagenet32 (both 1.2M samples), and CelebA (200K samples). The datasets were partitioned over 2-16 clients horizontally. We compared FedPCs to EiNets and Pyjuice and ran all methods with 5 different seeds.

To evaluate FCs in FL scenarios, we selected three tabular datasets that cover various application domains and data regimes present in the real world: one credit fraud dataset ($\sim 300$K samples), a medical dataset (breast cancer detection; $<1000$ samples), and the popular Income dataset ($>1$M samples). The selected datasets for FL cover low-data, medium-data, and large-data regimes (see App.~\ref{app:exp_details} for more details). Both balanced (breast cancer) and imbalanced (income, credit) datasets are included in our evaluation. We selected tabular datasets as they are well suited to investigate FCs in horizontal, vertical, and hybrid settings and represent various real-world applications. We compare FCs to multiple strong and widely used baselines. As a neural network architecture parameterization, we use TabNet~\citep{arik2020tabnetattentiveinterpretabletabular} which is tailored to tabular datasets. We train the networks with the widely used FedAvg (horizontal FL) and SplitNN (vertical FL) frameworks. Additionally, we compare FCs to FedTree~\citep{li2023fedtree} since tree models excel at tabular datasets. For details, see App. \ref{app:exp_details}.

\textbf{\textbf{(Q1)} FedPCs learn joint distributions over 
partitioned data in less time.} First, we validate that FedPCs
correctly and efficiently perform density estimation on partitioned datasets 
distributed over multiple clients. To this end, multiple datasets were distributed over a set of clients corresponding to horizontal (5 clients),
vertical (2 clients), and hybrid FL (2 clients). To demonstrate that FedPCs are also robust against label shifts, a common regime in FL, each client received data from only a subset of classes in the horizontal case, and local PCs were learned over the client samples. 
In the vertical case, we split data s.t. feature spaces of clients 
are disjoint, but
each client 
holds the same samples. In hybrid settings, data was distributed s.t. both feature- and sample-spaces among 
clients have overlaps (but no full overlap).
For all tabular datasets, the leaves of the FedPC were parameterized with MSPNs~\citep{molina2018MSPNs}, a member of the 
PC model family capable of performing density estimation on mixed data domains (i.e., continuous and discrete random variables). 
We chose MSPNs as the centralized models, which were learned using \textsc{learnSPN}, a recursive greedy structure learning algorithm for SPNs~\cite{gens2013LearnSPN}. For MNIST, EiNets with Gaussian densities were used as PC instantiations in all settings. Note that FedPCs were chosen to approximately match the size of centralized models, i.e., no model upscaling was performed.

Tab. \ref{tab:likleihoods} compares log-likelihoods and relative runtime of centralized PC training on the full datasets with log-likelihood scores and relative runtimes achieved by FedPC in different FL settings. FedPCs achieve the same log-likelihoods as centralized PCs on tabular datasets while being tremendously faster in training. Hence, we answer \textbf{(Q1)} affirmatively.

\begin{figure*}[t]
    \centering
    \includegraphics[width=.85\textwidth]{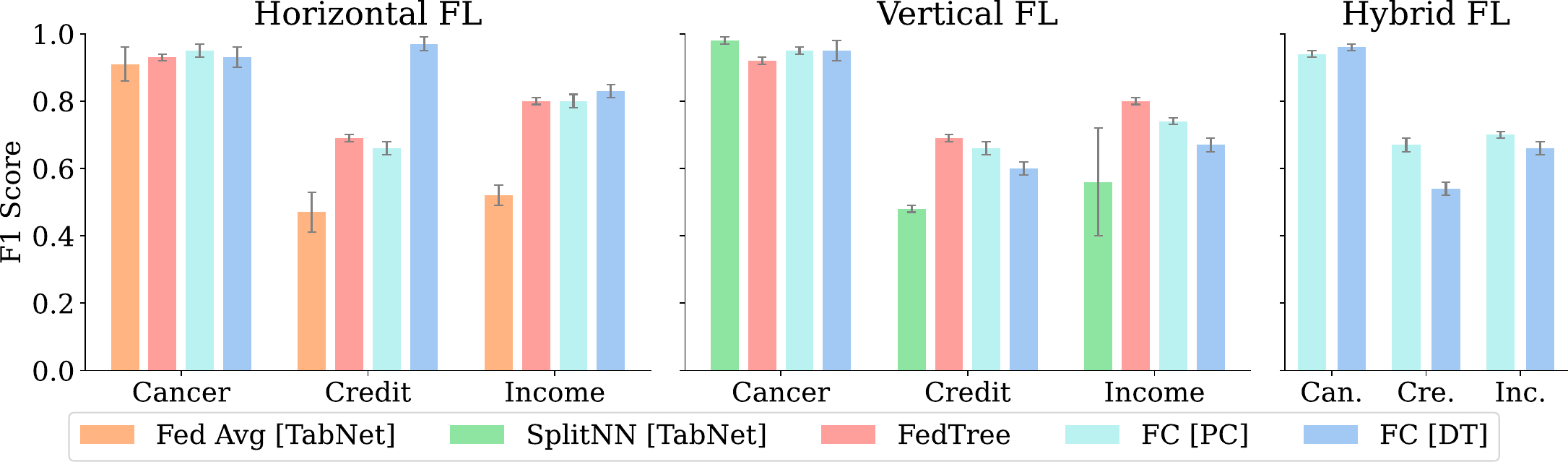}
    \caption{\textbf{FCs are competitive to prominent FL methods in all settings.} FCs achieve competitive performance on various classification tasks compared to prominent horizontal/vertical FL baselines. FCs also handle the more challenging setting of hybrid FL without performance drops. We reported the F1 score (higher is better).}
    \label{fig:fl_results}
    \vspace{-0.25cm}
\end{figure*}

\textbf{\textbf{(Q2)} FedPCs effectively scale up PCs.} To examine whether FedPCs can be leveraged to scale up PCs effectively, we trained an EiNet, PyJuice, and FedPC on CelebA, Imagenet32, and Imagenet. All models used the Poon-Domingos (PD) architecture. FedPCs were parameterized with EiNets, and data was distributed among 2, 4, 8, and 16 clients. The FedPC model and baseline models (EiNets and PyJuice) were selected to ensure that each fits within a single GPU (see App. \ref{app:exp_details} for system details). Einets and FedPCs were parameterized with Gaussian leave distributions, while PyJuice models were parameterized with Categorical distributions. The parameterizations were chosen based on empirical observations; for Einets and FedPCs, Gaussians worked best, while PyJuice Categoricals worked best.

For FedPC training, the images were distributed horizontally at random, s.t. each client holds approximately equally large subsets. The leaves and all baselines were trained with EM.
In Tab. \ref{tab:imageLLs}, we show nats normalized over samples and dimensions achieved by EiNets, PyJuice, and FedPC on the test set. 
It can be seen that with an increasing number of participating clients and, thus, a larger model, the density estimation performance also increases on all three datasets. We posit that this is because larger models exhibit higher expressivity, allowing them to capture statistical characteristics of the data better than smaller models. Also, higher nats scores achieved on the test set by larger models indicate that no overfitting appeared due to more model parameters. However, note that more exhaustive scaling will likely lead to overfitting. Finding the optimal model size/number of clients in a principled way is beyond the scope of this work and is left for future endeavors.
Besides better modeling performance, a larger number of clients reduces training time significantly (see Fig. \ref{fig:runtime}). FedPCs thus efficiently scale tractable probabilistic models to large datasets.

\textbf{\textbf{(Q3)} FCs achieve state of the art classification results in FL.}
FCs can be parameterized with different models in the leaves. We examine two parameterizations to solve a federated classification task on three tabular datasets. First, we use the FedPC (FC [PC]) from \textbf{(Q1)}, which can be used to solve discriminative tasks leveraging tractable computation of conditionals in PCs. The second FC parameterization we examine is decision trees (FC [DT]), representing an instantiation of a bagging model.
To see how FCs perform in federated classification tasks, we compare FCs to well-known methods for horizontal FL and vertical FL. The experiments were conducted on tabular datasets covering various real-world application domains and distribution properties. 
We employ TabNet and FedTree as strong baselines. In the horizontal FL setting, TabNet was trained using FedAvg; in the vertical FL setting, it was trained in a SplitNN fashion~\citep{Cellabos2020SplitNN}. The results were compared against our one-pass training. FCs yield comparable or even better results than the selected baselines on all datasets (see Fig \ref{fig:fl_results}; App. \ref{app:results}) while being significantly more flexible compared to the baselines. 

\textbf{\textbf{(Q4)} One-pass training retains performance.} To see how the proposed one-pass training compares to training PCs
with standard optimization algorithms such as EM, we define an FL setup where data exchange is allowed. 
This is necessary as we have to train the PC and FedPC architecture with EM to compare to our one-pass procedure. We used RAT-SPNs~\citep{peharz2020random} as leaf
parameterizations of the FedPC. Then, we trained a FedPC using standard EM (i.e., data exchange was allowed) and another FedPC with the same FedPC architecture on a vertically split dataset using our one-pass procedure. 
We report the final average log-likelihood of the test dataset, both for EM training 
and one-pass training (see Tab. \ref{tab:e2evs2s}). It can be seen that there is no significant decrease in log-likelihood in any case. 
Hence, our results indicate that one-pass training is preferable since it yields comparable model performance while being more communication efficient. 
\begin{table}
    \centering
    \begin{tabular}{c|cc}
    \multicolumn{1}{l|}{} & EM & one-pass \\ \hline
    Synth. Data  & $-53.6$ \scriptsize{$\pm 1.3$} & $-53.2$ \scriptsize{$\pm 1.2$}  \\
    Income  &  $-18.5$ \scriptsize{$\pm 0.1$}        & $-18.0$ \scriptsize{$\pm 0.5$} \\
    Breast-Cancer & $-52.3$ \scriptsize{$\pm 0.2$} & $-55.7$ \scriptsize{$\pm 0.2$} \\
    Credit & $-26.7$ \scriptsize{$\pm 1.2$} & $-28.3$ \scriptsize{$\pm 0.4$}
    \end{tabular}
    \caption{\textbf{One-pass training retains performance.} We trained the same FedPC architecture on various datasets using EM and one-pass training in a vertical setting. The average log-likelihood value of the hold-out test set across 10 runs is reported. 
    }
    \label{tab:e2evs2s}
    \vspace{-0.7cm}
\end{table}

\section{Conclusion}
In this work, we introduced federated circuits that hinge on an inherent connection between PCs and FL. We demonstrated that both the training speed and expressivity of PCs can be increased by learning PCs on scale across partitioned data.
Since our framework allows for the integration of various types of density estimators, other models and advances of PCs and other fields can be integrated seamlessly, maintaining the relevance of the federated approach for scaling.

\textbf{Limitations and Future Work.}
While our experiments showed that scaling PCs can considerably improve training speed and performance, scaling to such large-scale models requires sufficient computational resources.
For future work, investigating other parametrizations for FCs beyond PCs is promising.
Additionally, it is interesting how the probabilistic framework for hybrid FL could also benefit more traditional FL applications, apart from scaling PCs.

\section*{Acknowledgements}
This work is supported by the Hessian Ministry of Higher Education, Research, Science and the Arts (HMWK; projects “The Third Wave of AI”). Further, this work was supported
from the National High-Performance Computing project for Computational Engineering Sciences (NHR4CES). 

The Eindhoven University of Technology authors received support from their Department of Mathematics and Computer Science and the Eindhoven Artificial Intelligence Systems Institute.

\bibliography{main}

\appendix
\newpage
\onecolumn

\section{Discussion on Assumptions}
\label{app:assumptions}
As a preliminary to FCs, we introduced two assumptions that allowed us to construct the FC framework. Here, we provide some more background on these assumptions. For clarity, let us state the assumptions again.

\textbf{Assumption 1} (Mixture Marginals).
    There exists a joint distribution $p$ such that the relation $\int_{\mathbf{X} \setminus \mathbf{X}_S} p(x) = \sum_{l \in L} q(L=l) \cdot p_{S}(x | L=l)$ holds for all $x \in \mathcal{X}$. Here, $\mathbf{X}_{S} \subseteq \mathbf{X}$ is a subset of the union of client random variables $\mathbf{X} = \cup_{c \in \mathcal{C}} \mathbf{X}_c$. Further, $\mathcal{X} = \bigtimes_{c \in \mathcal{C}} \mathcal{X}_c$ is the support of $\mathbf{X}$, each $p_{S}$ is defined over $\mathbf{X}_S \subseteq \mathbf{X}$ and $q$ is a prior over a latent $L$.

\textbf{Assumption 2} (Cluster Independence).
    Given disjoint sets of random variables $\mathbf{X}_1, \cdots, \mathbf{X}_n$ and a joint distribution $p(\mathbf{X}_1, \cdots, \mathbf{X}_n)$, assume that a latent $L$ can be introduced s.t. the joint can be represented as $p(\mathbf{X}_1, \cdots, \mathbf{X}_n) = \sum_l q(L=l) \prod_{i=1}^n p(\mathbf{X}_i | L=l)$ where $q$ is a prior distribution over the latent $L$.

As discussed in the main paper, Assumption \ref{assum:decomposition} ensures that the data that resides on all participating clients is sufficient to learn $p(\mathbf{X})$, at least in the limit of infinite samples available. However, this only covers the federated learning perspective of this assumption. There is also a PC perspective on this assumption. For this, let us introduce the induced tree representation of PCs from~\citep{zhaoa16CollapsedVarInf}:

\begin{defin}
    \label{def:induced_pc}
        Induced Trees~\citep{zhaoa16CollapsedVarInf}. Given a complete and decomposable PC $s$ over $\mathbf{X} = \{X_1, \dots, X_n\}$, $\mathcal{T} = (\mathcal{T}_V, \mathcal{T}_E)$ is called an induced tree PC from $s$ if
    \begin{enumerate}
        \item $\Node \in \mathcal{T}_V$ where $\Node$ is the root of $s$.
        \item for all sum nodes $\SumNode \in \mathcal{T}_V$, exactly one child of $\SumNode$ in $s$ is in $\mathcal{T}_V$, and the corresponding edge is in $\mathcal{T}_E$.
        \item for all product node $\ProductNode \in \mathcal{T}_V$, all children of $\ProductNode$ in $s$ are in $\mathcal{T}_V$, and the corresponding edges in $\mathcal{T}_E$.
    \end{enumerate}
\end{defin}

We can use Def. \ref{def:induced_pc} to represent decomposable and complete PCs as mixtures~\citep{zhaoa16CollapsedVarInf}.

\begin{prop}[Induced Tree Representation]
    \label{prop:induced_tree}
        Let $\tau_s$ be the total number of induced trees in $s$. Then the output at the root of $s$ can be written as $\sum_{t=1}^{\tau_s} \prod_{(k, j) \in \mathcal{T}_{t E}} w_{k j} \prod_{i=1}^n p_t(X_i = x_i)$, where $\mathcal{T}_t$ is the $t$-th unique induced tree of $s$ and $p_t(X_i)$ is a univariate distribution over $X_i$ in $\mathcal{T}_t$ as a leaf node.
\end{prop}

Using Prop. \ref{prop:induced_tree}, we see that any decomposable and smooth PC can be represented as a mixture without any hierarchy, i.e., we can collapse the PC structure into a structure of depth one. Since marginalizing over a decomposable and smooth PC yields another  decomposable and smooth PC again, and since the marginalized PC can be represented as an induced tree, Assumption \ref{assum:decomposition} is a standard assumption in the PC literature.

Also, Assumption \ref{assum:cluster_independence} can be viewed from a PC perspective. In popular structure learning algorithms such as LearnSPN~\cite {gens2013LearnSPN}, a PC is learned by alternating data clustering with testing for independent subsets of features. Thus, the ultimate goal of algorithms like LearnSPN is to find clusters in which subsets of random variables are considered independent in order to maximize log-likelihood. Therefore, Assumption \ref{assum:cluster_independence} is closely related to LearnSPN and, thus, a common assumption in PC modeling.

\section{Proofs}
\label{app:proofs}
In this section we give full proofs for our propositions in the paper.

\subsection{FedPCs and Principle of Maximum Entropy}
\label{app:max_entropy}
Assumption 2 aligns with the principle of maximum entropy: we aim to find the joint distribution with maximum entropy \textit{within} clusters while allowing for dependencies among clients’ random variables and ensuring the marginals for each client are preserved. Although multiple joint distributions can preserve the marginals, non-maximal entropy solutions introduce additional assumptions or prior knowledge, limiting flexibility. By assuming independence of all variables within a cluster, we efficiently construct the maximum entropy distribution via a mixture of product distributions.
For independent variables, the product distribution maximizes entropy, as can be shown by leveraging the joint and conditional differential entropy.
Given random variables $\mathbf{X} = X_1, \dots, X_n$ and a density $p$ defined over support $\mathcal{X} = \mathcal{X}_1 \times \cdots \times \mathcal{X}_n$, the joint differential entropy is defined as:
\begin{equation}
    h(\mathbf{X}) = \int_{\mathcal{X}} p(x_1, \dots, x_n) \, \text{log} p(x_1, \dots, x_n)
\end{equation}
The conditional differential entropy for two sets of random variables $\mathbf{X}$ and $\mathbf{Y}$ and a joint distribution $p(\mathbf{X}, \mathbf{Y})$ defined over support $\mathcal{X} \times \mathcal{Y}$ is defined analogously: 
\begin{equation}
    h(\mathbf{X} | \mathbf{Y}) = \int_{\mathcal{X}, \mathcal{Y}} p(\mathbf{x}, \mathbf{y}) \, \text{log}p(\mathbf{x} | \mathbf{y})
\end{equation}
Given two sets of random variables $\mathbf{X}$, $\mathbf{Y}$ with densities $p(\mathbf{X})$ and $p(\mathbf{Y})$ and support $\mathcal{X}$, $\mathcal{Y}$ respectively, the joint $p(\mathbf{X}, \mathbf{Y}) = p(\mathbf{X}) \cdot p(\mathbf{Y})$ is the maximum entropy distribution if $\mathbf{X}$ and $\mathbf{Y}$ are mutually independent.
\begin{proof}
 We consider the two cases that $\mathbf{X}$ and $\mathbf{Y}$ are mutually independent and that they are not mutually independent. The joint entropy can be written as $h(\mathbf{X}, \mathbf{Y}) = h(\mathbf{X} | \mathbf{Y}) + h(\mathbf{Y})$. In the case of mutual independence, this reduces to $h(\mathbf{X}, \mathbf{Y}) = h(\mathbf{X}) + h(\mathbf{Y})$. Hence it has to be shown that $h(\mathbf{X} | \mathbf{Y}) < h(\mathbf{X})$ holds if $\mathbf{X}$ and $\mathbf{Y}$ are not mutually independent:
 \begin{align*}
     & h(\mathbf{X} | \mathbf{Y}) < h(\mathbf{X}) \\
     \equiv & - \int_{\mathcal{X}, \mathcal{Y}} p(\mathbf{x}, \mathbf{y}) \text{log} p(\mathbf{x} | \mathbf{y}) < - \int_{\mathcal{X}, \mathcal{Y}} p(\mathbf{x}, \mathbf{y}) \text{log} p(\mathbf{x}) \\
     \equiv & - \bigg( \int_{\mathcal{X}, \mathcal{Y}} p(\mathbf{x}, \mathbf{y}) \text{log} p(\mathbf{x} | \mathbf{y}) - \int_{\mathcal{X}, \mathcal{Y}} p(\mathbf{x}, \mathbf{y}) \text{log} p(\mathbf{x}) \bigg) < 0 \\
     \equiv & - \bigg( \int_{\mathcal{X}, \mathcal{Y}} p(\mathbf{x}, \mathbf{y}) \text{log} \frac{p(\mathbf{x} | \mathbf{y})}{p(\mathbf{x})} \bigg) < 0
 \end{align*}
Since $\mathbf{X} \indep \mathbf{Y}$ holds where $\indep$ means mutual independence, $\frac{p(\mathbf{x} | \mathbf{y})}{p(\mathbf{x})} \neq 1$ at least for some $\mathbf{x}, \mathbf{y}$. Since the mutual independence $I(\mathbf{X}, \mathbf{Y}) = \int_{\mathcal{X}, \mathcal{Y}} p(\mathbf{x}, \mathbf{y}) \text{log} \frac{p(\mathbf{x}, \mathbf{y})}{p(\mathbf{x}) \cdot p(\mathbf{y})}$ can be represented as $I(\mathbf{X}, \mathbf{Y}) = h(\mathbf{X}) - h(\mathbf{X} | \mathbf{Y})$, $I(\mathbf{X}, \mathbf{Y}) \geq 0$ holds and $- \Big( \int_{\mathcal{X}, \mathcal{Y}} p(\mathbf{x}, \mathbf{y}) \text{log} \frac{p(\mathbf{x} | \mathbf{y})}{p(\mathbf{x})} \Big) = h(\mathbf{X} | \mathbf{Y}) - h(\mathbf{X})$ it follows that $h(\mathbf{X}) > h(\mathbf{X} | \mathbf{Y})$.
 
\end{proof}

\newpage
\clearpage
\section{Communication Efficiency}
\label{app:commeff}
\begin{figure}[h!]
    \centering
    \includegraphics[scale=0.4]{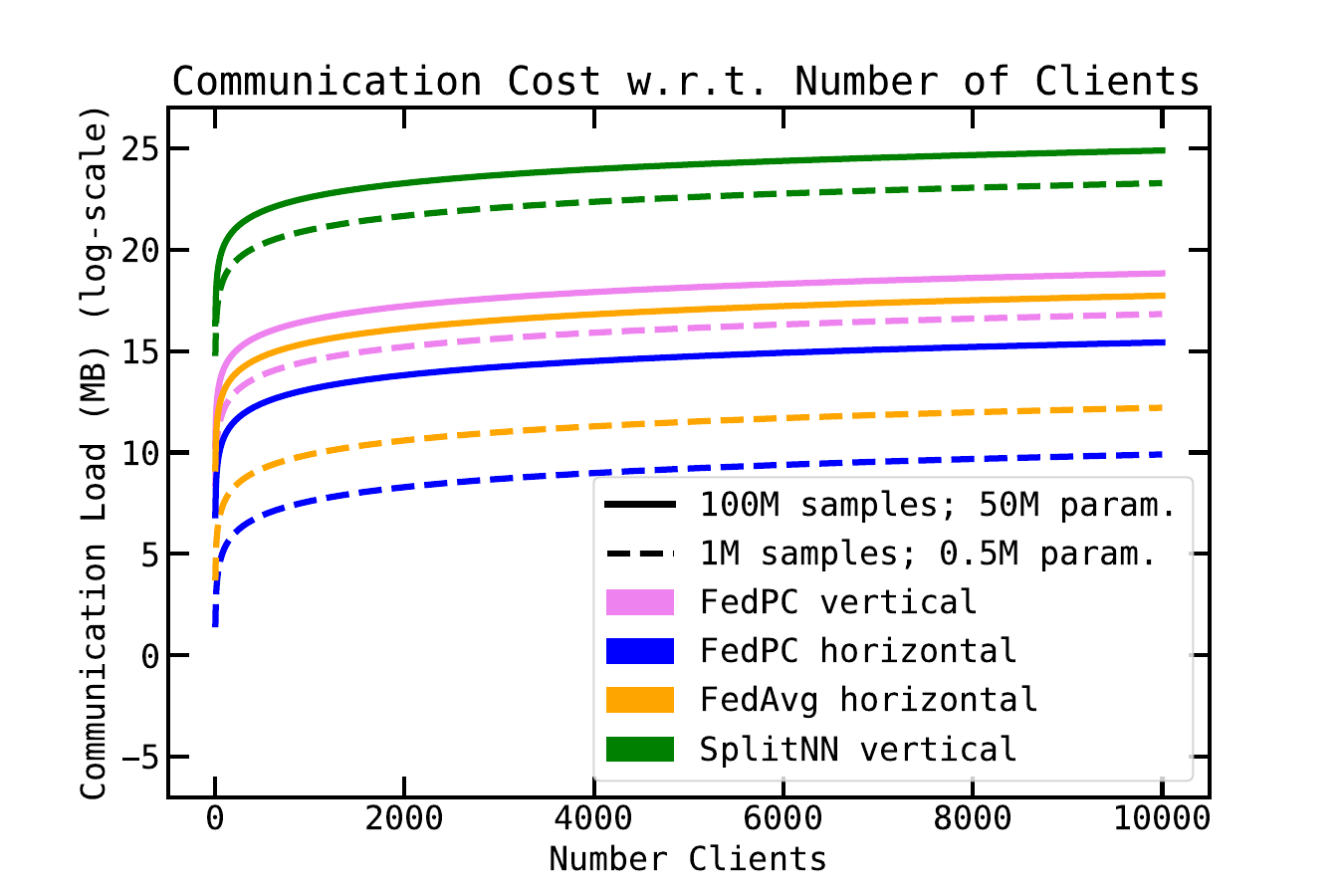}
    \caption{\textbf{FedPCs are communication-efficient.} We compare communication cost in Megabytes (MB) sent over the network during one full training of a model (0.5M/50M parameters) on a dataset (1M/100M samples) using results from Section 3.4. Results are shown on log-scale. It can be seen that FedPCs significantly reduce communication cost of training.}
    \label{fig:comm_effic}
\end{figure}

Communication efficiency is a critical property when it comes to learning models across multiple machines, as it is done in FL. Here, in addition to our theoretical results, we more intuitively provide further details on the communication efficiency of FCs. For that, we plot the communication cost in Megabytes (MB) required to train a FedPC vs. FedAvg/SplitNN in horizontal/vertical FL settings with datasets of different sizes (1M and 100M samples). Regardless of the number of samples in the dataset, FedPCs are more communication efficient compared to our baselines in both horizontal and vertical settings (see Fig. \ref{fig:comm_effic}).

\section{Experimental Details}
\label{app:exp_details}
\subsection{Datasets}
The following describes the datasets used in our experiments. If not stated differently, the datasets were distributed across clients as follows: 

In horizontal cases, we either split samples randomly across clients (done for all binary classification tasks) or we distribute a subset of the dataset corresponding to a certain label (e.g. the 0 in MNIST) to one client. 

In vertical cases, we split tabular datasets randomly along the feature-dimension, i.e. each client gets all samples but a random subset of features assigned. For image data, we split the images into non-overlapping patches which were then distributed to the clients.

In hybrid cases, we split tabular datasets along both, the feature and the sample-dimension. We do this s.t. at least two clients have at least one randomly chosen feature in commeon (but hold different samples thereof). For image data, we split images into overlapping patches, sample a subset of the dataset and assign the resulting subsets to clients.

\textbf{Income Dataset.}
We used the Income dataset from \url{https://www.kaggle.com/datasets/wenruliu/adult-income-dataset}. This dataset represents a binary classification problem with 14 features and approximate 450K samples in the train and 900 samples in the test set. We encoded discrete variables to numerical values using TargetEncoder from sklearn. Additionally, missing values were imputed using the median of the corresponding feature. Further we standardized all features.

\textbf{Breast Cancer Dataset.}
We used the Breast Cancer dataset from \url{https://www.kaggle.com/datasets/uciml/breast-cancer-wisconsin-data}. It represents a binary classification problem with 31 features and 570 samples. We split the dataset into 450 training samples and 120 test samples. We standardized all features for training.

\textbf{Credit Dataset.}
We used the Give Me Some Credit dataset from \url{https://www.kaggle.com/c/GiveMeSomeCredit}. The dataset represents a binary classification task with 10 features, 1.5M training samples and 100K test samples. We encoded discrete variables to numerical values using TargetEncoder from sklearn. Additionally, missing values were imputed using the median of the corresponding feature. Further we standardized all features.

\textbf{MNIST.}
We used the MNIST dataset provided by pytorch. It contains 70K hand-written digits between 0 and 9 as 28x28 images (60K train, 10K test). We standardized all features as preprocessing.

\textbf{Imagenet/Imagenet32.}
We used the Imagenet dataset provided by pytorch. It consists of about 1.2M images showing objects of 1000 classes. The images come in different resolutions; we resized each image to 64x64 (Imagenet) and 32x32 (Imagenet32) pixels, applied center cropping, and standardized all features as preprocessing. We distributed samples randomly across clients as a simple dataset partioning.

\subsection{Discretization}
In our experimental setup, FCs and Einets were parameterized with Gaussian leaves and fitted on RGB image data. Since image data is discrete (takes integer values from 0-255) and Gaussians are defined over a continuous domain and thus define a probability \textit{density} rather than a probability \textit{mass} function, we have to discretize the Gaussian leaves to obtain the probability for a given image $\mathbf{x}$. Therefore, we construct 255 buckets, discretizing a Gaussian with parameters $\mu$ and $\sigma$ by computing the probability mass as $p(x) = \Phi(\frac{x - \mu + \frac{1}{255}}{\sigma}) - \Phi(\frac{x - \mu}{\sigma})$. The computation graph will remain fixed since the probabilistic semantics of PCs hold for densities and probability mass functions.

\subsection{Training \& Hyperparameters}
\textbf{PyJuice.} For PyJuice we follow the training setup of~\citep{liu2024scalingtractableprobabilisticcircuits} and train the models on randomly cropped 32x32 patches. For Imagenet32, this corresponds to the full image, while for CelebA and Imagenet, this corresponds to a random 32x32 block of the image. We used EM as an optimization procedure.

\textbf{FCs and Einets.} For FCs and Einets, we followed the setup from~\citep{perharz2020einsum} and trained all models on the full image (i.e., 32x32 for Imagenet32 and 64x64 for Celebi/Imagenet). We used EM as an optimization procedure.

We ran all experiments with 5 different seeds (0-4). The following tables show the setting of all relevant hyperparameters for each dataset and FL setting.
\begin{table}[h!]
\centering
\begin{tabular}{c|ccccc}
FL-Setting                  & Dataset & Structure & Threshold & min\_num\_instances & glueing       \\ \hline
\multirow{3}{*}{horizontal} & Income  & learned   & 0.3       & 200                 & -             \\
                            & Credit  & learned   & 0.5       & 200                 & -             \\
                            & Cancer  & learned   & 0.4       & 300                 & -             \\
\multirow{3}{*}{vertical}   & Income  & learned   & 0.4       & 100                 & combinatorial \\
                            & Credit  & learned   & 0.5       & 50                  & combinatorial \\
                            & Cancer  & learned   & 0.4       & 300                 & combinatorial \\
\multirow{3}{*}{hybrid}     & Income  & learned   & 0.4       & 100                 & combinatorial \\
                            & Credit  & learned   & 0.5       & 50                  & combinatorial \\
                            & Cancer  & learned   & 0.4       & 300                 & combinatorial
\end{tabular}
\caption{Hyperparameters used in our experiments for all tabular datasets.}
\end{table}

\begin{table}[h!]
\centering
\begin{tabular}{c|cccc}
                      & MNIST             & Imagenet32   & Imagenet         & CelebA\\ \hline
num\_epochs           & 5                 & 10             & 10     & 10\\
batch\_size           & 64                & 64             & 64     & 64\\
online\_em\_frequency & 5                 & 10             & 50     & 10\\
online\_em\_stepsize  & 0.1               & 0.25           & 0.5      & 0.25\\
Structure             & poon-domingos     & poon-domingos  & poon-domingos    & poon-domingos\\
pd\_num\_pieces       & 8                 & 8              & 8     & 8\\
K                     & 10                & 40             & 40     & 40\\
Leaf Distribution     & Gaussian          & Gaussian       & Gaussian     & Gaussian\\
min\_var              & $1 \cdot 10^{-3}$ & $1 \cdot 10^{-3}$ & $1 \cdot 10^{-3}$   & $1 \cdot 10^{-3}$\\
max\_var              & $0.25$ & $0.25$  & $0.25$ & $0.25$
\end{tabular}
\caption{Hyperparameters used in our experiments for image datasets.}
\end{table}

\subsection{Hardware}
All experiments were conducted on Nvidia DGX machines with Nvidia A100 (40GB) GPUs, AMD EPYC 7742 64-Core Processor and 2TiB of RAM.

\section{Further Results}
\label{app:results}
Here, we provide further experimental details on FCs.

\textbf{FL Classification Results.} We compare FCs to several baselines in horizontal, vertical, and hybrid FL. In horizontal FL, we compare against FedAvg (using TabNet~\citep{arik2020tabnetattentiveinterpretabletabular}) and FedTree~\citep{li2023fedtree}; in vertical FL, we compare against SplitNN (also using TabNet) and FedTree. In hybrid FL, we compare different parameterizations of FCs (FedPCs and FCs parameterized with decision trees). We find that FCs are competitive or outperforming the selected baselines in all FL settings (see Tab. \ref{tab:full_FL_results}). This makes them a very flexible FL framework that still yields high-performing models.

\begin{table}[h!]
\resizebox{\textwidth}{!}{
\begin{tabular}{c|c|cccccc}
&                             & \multicolumn{2}{c}{Cancer}        & \multicolumn{2}{c}{Credit}        & \multicolumn{2}{c}{Income}        \\ \hline
&                             & Acc.            & F1              & Acc.            & F1              & Acc.            & F1              \\
\multirow{9}{*}{\rotatebox{90}{Horizontal FL}} & FedAvg {[}TabNet{]} (5 cl.)  & $0.92 \pm 0.03$ & $0.92 \pm 0.03$ & $0.71 \pm 0.11$ & $0.48 \pm 0.04$ & $0.68 \pm 0.06$ & $0.51 \pm 0.03$ \\
& FedAvg {[}TabNet{]} (10 cl.) & $0.92 \pm 0.04$ & $0.91 \pm 0.05$ & $0.56 \pm 0.12$ & $0.47 \pm 0.06$ & $0.64 \pm 0.06$ & $0.52 \pm 0.03$ \\
& FedTree (5 cl.)              & $0.93 \pm 0.01$ & $0.92 \pm 0.01$ & $0.91 \pm 0.01$ & $0.63 \pm 0.01$ & $0.88 \pm 0.01$ & $0.82 \pm 0.02$ \\
& FedTree (10 cl.)             & $0.94 \pm 0.01$ & $0.93 \pm 0.01$ & $0.92 \pm 0.01$ & $0.69 \pm 0.01$ & $0.87 \pm 0.01$ & $0.80 \pm 0.01$ \\
& FC {[}PC{]} (5 cl.)          & $0.98 \pm 0.01$ & $0.98 \pm 0.01$ & $0.93 \pm 0.02$ & $0.68 \pm 0.02$ & $0.87 \pm 0.02$ & $0.80 \pm 0.01$ \\
& FC {[}PC{]} (10 cl.)         & $0.95 \pm 0.02$ & $0.95 \pm 0.02$ & $0.93 \pm 0.01$ & $0.66 \pm 0.02$ & $0.87 \pm 0.01$ & $0.80 \pm 0.02$ \\
& FC {[}DT{]} (5 cl.)          & $0.95 \pm 0.03$ & $0.93 \pm 0.02$ & $0.92 \pm 0.01$ & $0.67 \pm 0.01$ & $0.89 \pm 0.01$ & $0.83 \pm 0.01$ \\
& FC {[}DT{]} (10 cl.)         & $0.95 \pm 0.02$ & $0.93 \pm 0.03$ & $0.92 \pm 0.01$ & $0.97 \pm 0.02$ & $0.89 \pm 0.01$ & $0.83 \pm 0.02$ \\
& SplitNN {[}TabNet{]}         & -               & -               & -               & -               & -               & -              \\ \hline
\multirow{9}{*}{\rotatebox{90}{Vertical FL}} & SplitNN {[}TabNet{]} (2 cl.) & $0.98 \pm 0.01$ & $0.98 \pm 0.01$ & $0.93 \pm 0.01$ & $0.48 \pm 0.01$ & $0.56 \pm 0.25$ & $0.42 \pm 0.17$ \\
& SplitNN {[}TabNet{]} (3 cl.) & $0.98 \pm 0.01$ & $0.98 \pm 0.01$ & $0.93 \pm 0.01$ & $0.48 \pm 0.01$ & $0.62 \pm 0.20$ & $0.56 \pm 0.16$ \\
& FedTree (2 cl.)              & $0.94 \pm 0.01$ & $0.93 \pm 0.01$ & $0.92 \pm 0.01$ & $0.69 \pm 0.02$ & $0.87 \pm 0.01$ & $0.80 \pm 0.01$ \\
& FedTree (3 cl.)              & $0.93 \pm 0.01$ & $0.92 \pm 0.01$ & $0.92 \pm 0.01$ & $0.69 \pm 0.01$ & $0.87 \pm 0.01$ & $0.80 \pm 0.01$ \\
& FC {[}PC{]} (2 cl.)          & $0.96 \pm 0.01$ & $0.96 \pm 0.01$ & $0.92 \pm 0.01$ & $0.67 \pm 0.01$ & $0.84 \pm 0.02$ & $0.74 \pm 0.01$ \\
& FC {[}PC{]} (3 cl.)          & $0.95 \pm 0.01$ & $0.95 \pm 0.01$ & $0.92 \pm 0.01$ & $0.66 \pm 0.02$ & $0.84 \pm 0.01$ & $0.74 \pm 0.01$ \\
& FC {[}DT{]} (2 cl.)          & $0.96 \pm 0.01$ & $0.96 \pm 0.02$ & $0.93 \pm 0.01$ & $0.60 \pm 0.02$ & $0.83 \pm 0.02$ & $0.67 \pm 0.02$ \\
& FC {[}DT{]} (3 cl.)          & $0.95 \pm 0.01$ & $0.95 \pm 0.03$ & $0.93 \pm 0.01$ & $0.60 \pm 0.02$ & $0.82 \pm 0.02$ & $0.67 \pm 0.02$ \\
& FedAvg {[}TabNet{]}          & -               & -               & -               & -               & -               & -              \\ \hline
\multirow{7}{*}{\rotatebox{90}{Hybrid FL}} & FC {[}PC{]} (2 cl.)  & $0.94 \pm 0.01$ & $0.94 \pm 0.01$ & $0.92 \pm 0.01$ & $0.67 \pm 0.01$ & $0.82 \pm 0.02$ & $0.71 \pm 0.01$ \\
& FC {[}PC{]} (3 cl.)  & $0.94 \pm 0.01$ & $0.94 \pm 0.01$ & $0.92 \pm 0.01$ & $0.67 \pm 0.02$ & $0.80 \pm 0.01$ & $0.70 \pm 0.01$ \\
& FC {[}DT{]} (2 cl.)  & $0.96 \pm 0.01$ & $0.96 \pm 0.02$ & $0.93 \pm 0.01$ & $0.60 \pm 0.02$ & $0.82 \pm 0.02$ & $0.66 \pm 0.02$ \\
& FC {[}DT{]} (3 cl.)  & $0.96 \pm 0.01$ & $0.96 \pm 0.01$ & $0.93 \pm 0.01$ & $0.54 \pm 0.02$ & $0.82 \pm 0.02$ & $0.66 \pm 0.02$ \\
& FedAvg {[}TabNet{]}  & -               & -               & -               & -               & -               & -               \\
& SplitNN {[}TabNet{]} & -               & -               & -               & -               & -               & -               \\
& FedTree              & -               & -               & -               & -               & -               & -        
\end{tabular}
}
\caption{\textbf{All Classification results of FL experiments.} Here, we show the detailed performances of FC, FedAvg, and SplitNN in all three FL settings. It can be seen that FCs, while being much more flexible than our baselines, still achieve competitive or better results on various classification tasks.}
\label{tab:full_FL_results}
\end{table}
\end{document}